\documentclass[12pt]{article}

\usepackage{amsmath}
\usepackage{scicite}
\usepackage{times}
\usepackage{xcolor}
\usepackage{graphicx}
\usepackage{geometry}
\usepackage{float}
\usepackage{caption}
\usepackage{hyperref}

\newfloat{Movie}{tbp}{mvi}

\usepackage{layout}

\DeclareUnicodeCharacter{211D}{~}

\newenvironment{sciabstract}{%
\begin{quote} \bf}
{\end{quote}}

\title{RoboBallet: Planning for Multi-Robot Reaching with Graph Neural Networks and Reinforcement Learning}

\author
{Matthew Lai$^{1,2}$, Keegan Go$^{3}$, Zhibin Li$^{2}$, Torsten Kr\"{o}ger$^{3}$,\\
Stefan Schaal$^{3}$, Kelsey Allen$^{1\ast}$, Jonathan Scholz$^{1\ast}$ \\
\\
\normalsize{$^{1}$Google DeepMind, $^{2}$University College London}, $^{3}$Intrinsic\\
\normalsize{$^\ast$ These authors contributed equally to this work.}
}

\date{}

\newcommand{\figclearpage}{}
\renewcommand{\listoffigures}{}

\begin{document} 

\baselineskip24pt

\maketitle 

{\color{blue}
This is the author's version of the work. It is posted here by permission of the AAAS for personal use, not for redistribution. The definitive version was published in Science Robotics on 2025-09-03, DOI: 10.1126/scirobotics.ads1204. \href{https://doi.org/10.1126/scirobotics.ads1204}{[Final Published Version]}}

\begin{sciabstract}
Modern robotic manufacturing requires collision-free coordination of multiple robots to complete numerous tasks in shared, obstacle-rich workspaces. Although individual tasks may be simple in isolation, automated joint task allocation, scheduling, and motion planning under spatio-temporal constraints remain computationally intractable for classical methods at real-world scales. Existing multi-arm systems deployed in the industry rely on human intuition and experience to design feasible trajectories manually in a labor-intensive process. To address this challenge, we propose a reinforcement learning (RL) framework to achieve automated task and motion planning, tested in an obstacle-rich environment with eight robots performing 40 reaching tasks in a shared workspace, where any robot can perform any task in any order. Our approach builds on a graph neural network (GNN) policy trained via RL on procedurally-generated environments with diverse obstacle layouts, robot configurations, and task distributions. It employs a graph representation of scenes and a graph policy neural network trained through reinforcement learning to generate trajectories of multiple robots, jointly solving the sub-problems of task allocation, scheduling, and motion planning. Trained on large randomly generated task sets in simulation, our policy generalizes zero-shot to unseen settings with varying robot placements, obstacle geometries, and task poses. We further demonstrate that the high-speed capability of our solution enables its use in workcell layout optimization, improving solution times. The speed and scalability of our planner also open the door to new capabilities such as fault-tolerant planning and online perception-based re-planning, where rapid adaptation to dynamic task sets is required.

Summary: Training a graph neural network policy using reinforcement learning enables scalable task and motion planning in multi-robot reaching.

\end{sciabstract}

\section*{INTRODUCTION}

In robotics applications such as welding \cite{pellegrinelli2017multi}, assembly \cite{yamada1995development}, painting \cite{chen2008automated}, construction \cite{hartmann2022long}, and object manipulation \cite{xian2017closed}, robotic arms are programmed to perform many tasks in a shared workspace. To reduce the overall execution time and increase throughput, more robots can be placed in the same cell to perform tasks concurrently. Such high density setups have the potential for reduced execution time with minimal increase in cost and space, compared to duplicating entire workcells.

However, coordinating multiple robots efficiently in confined spaces is both theoretically and technically difficult because the planner must solve the task assignment, scheduling, and collision-free motion planning problems jointly, all of which have high combinatorial complexity, often in environments with complex obstacle geometries. This work addresses this challenge in the setting where multiple robots can perform all tasks in any order, which represents the most challenging scenario in terms of the size of the search space.

Finding a collision-free path in joint space that moves a robot end effector to a pose target is the motion planning aspect of the problem, and has been conventionally addressed using sampling-based approaches such as rapidly exploring random trees (RRT) and its variants\cite{lavalle1998rapidly}\cite{lavalle2001rapidly}. However, although these sampling-based algorithms are designed to tackle high dimensional configuration spaces and offer probabilistic completeness, their runtime scales exponentially in the dimensionality of the configuration space and the complexity of the obstacles\cite{karaman2011sampling}. In practice, these techniques are only suitable for relatively small configuration spaces (2 or fewer robots\cite{hartmann2022long}), when attempting to plan with all robotics at the same time. One potential solution is to plan for each robot sequentially, however, this formulation introduces its own challenges as other robots would have to be treated as moving obstacles, which sampling-based motion planners also struggle with, and would require techniques such as converting paths of other robots into swept volumes, which then unduly limits the search space of the planner and might lead to sub-optimal solutions or planning failures.

Similarly to the path planning problem, the task scheduling problem (to decide the order in which the tasks should be done) is also polynomial space (PSPACE)-Complete\cite{canny1988complexity}\cite{vega2020task}, formalized as the travelling salesmen problem (TSP). Even though exact solutions to the TSP are not computationally feasible for real world problem sizes, many approximate solutions exist with varying optimality / efficiency trade-offs. The main challenge in adapting these solutions to task scheduling in our context is that instead of having a fixed cost to move between each pair of tasks, as is the case in the classically formulated TSP, the joint-space trajectories required to move from task to task can be modified depending on the trajectories of other robots, and each potential trajectory has a different cost. Furthermore, there is often more than one inverse-kinematics (IK) solution per task, either due to the task not being fully constrained, or the robot having more degrees of freedom (DoF) than the task. In the case of 7-DoF robot arms, each 6-DoF task pose has up to an infinite number of IK solutions. A good path planning algorithm must take the entire sequence of tasks into account when selecting the best IK solutions to use to solve each task.

The task allocation problem adds yet another layer of complexity. Finding an allocation of tasks to robots that minimizes the total execution time is conceptually similar to a multi-container variant of the classical knapsack problem\cite{mathews1896partition}, except we do not know the marginal cost of allocating an additional task to a robot without solving the entire planning problem, nor are the costs independent of other tasks being solved by the robots. Therefore, algorithmic solutions to the knapsack problem are not practically helpful in this setting.

As a result of the high asymptotic complexity of each of the sub-problems, it has not been computationally feasible to combine existing algorithms with completeness guarantees to jointly solve high-density multi-robot planning on real world problem sizes, due to their exponential scaling in the number of robots, number of tasks, and complexity of obstacles\cite{ha2020learning}. Instead, existing approaches focus on breaking the problem down into manageable sub-problems, solving them one at a time, often iteratively, to generate solutions with realistic compute requirements \cite{hartmann2022long}. However, this approach sacrifices potential solution optimality for computational feasibility, and still does not scale to environment complexities of real world applications, as the current state of the art reported scalability to only five robots and 10 tasks, by iterating between a high level task planner and a low level motion planner, with some heuristics based on reachability and availability \cite{pan2021general}. By necessity, industry currently relies instead on manual trajectory planning with complex schedules of interlocks to ensure collision-free operation. These trajectories often take hundreds or thousands of hours to design by hand \cite{pellegrinelli2017multi}, and any change to the tasks or the environment requires a time-consuming manual replanning.

To tackle this challenge, we present RoboBallet, a learned heuristics-based approach for a task and motion planning on randomly-generated environments, capable of planning multi-arm reaching trajectories in environments different from those seen during training, with arbitrary obstacle geometries, task poses, and robot placements. Task and motion planning is a large class of problems that require both high-level discrete scheduling and low-level continuous planning to solve. In this work, we focus on the reaching subclass for simplicity. Specifically, we address problems where scalability and efficient task scheduling is the main challenge, with no implicit or explicit task inter-dependencies and where each individual task is simple to accomplish. For extensions to manipulation tasks or problems with ordering/compatibility constraints, see the Limitations and Future Work section.

Our approach tackles the task allocation, scheduling, and motion planning problems jointly with no hand-designed simplifications common in existing approaches, such as pre-sampling a fixed number of IK solutions, pre-allocating tasks to robots based on heuristics, or planning sub-trajectories for one robot at a time. Our key insight is to adopt a deep reinforcement learning (deep RL) approach which trains an \textbf{agent} to control the arms at each time-step. This shifts expensive online computations to an offline phase, amortizing and shifting the compute requirement from the planning phase to the training phase, benefiting applications that require planning solutions on many similar setups or a setup with many variations. By training on configuration variations that we wish to support, for example robot positioning and morphology, task types, or obstacle types, RoboBallet is able to quickly generate plans for a wide range of environments and novel configurations at test time without further training (Movie \ref{fig:train_eval_setup}). After training we can either simply execute these controllers in the real world, or for safety we can generate the solution trajectory by unrolling the policy step-by-step with a forward kinematic simulator, so it can be inspected and potentially processed further.

\begin{figure}[H]
\includegraphics[width=7.3in]{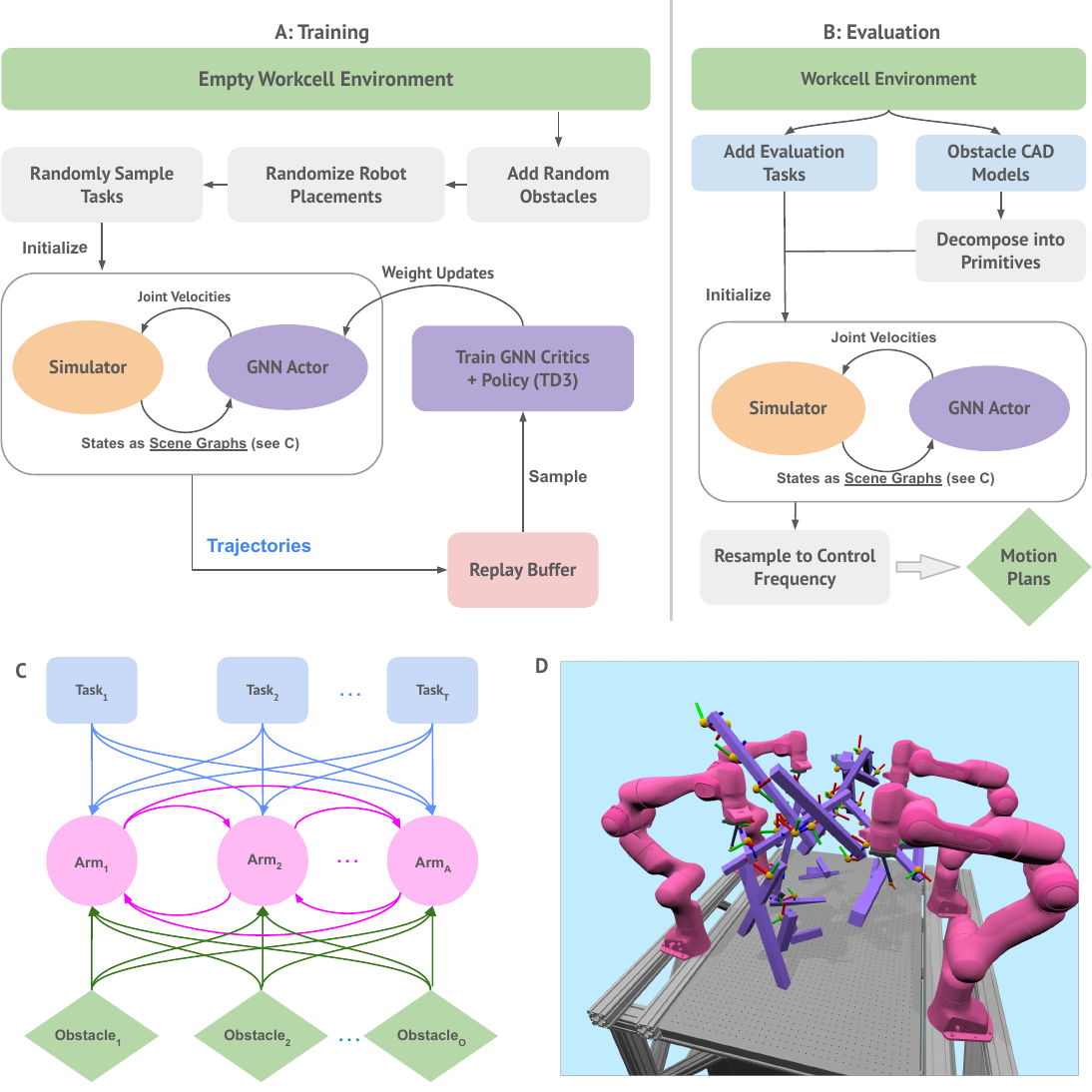}
\caption{\textbf{Training and evaluation pipeline.}\\
(\textbf{A}) The training pipeline. Each episode starts with an empty table, upon which we add random cuboid obstacles, randomize robot placements, and randomly sample end effector tasks. The GNN actor interacts with the kinematic simulator to try to complete the tasks. The trajectories are stored in a replay buffer and used to train a critic and policy function using TD3. (\textbf{B}) The evaluation pipeline, where externally-defined tasks, together with obstacles decomposed into collision primitives, are used to construct the simulated environment, with which the GNN interacts in the same way as in training. (\textbf{C}) A topographical view of the scene graph, containing robot, task, and obstacle nodes, with bi-directional edges between robots, and uni-directional edges from each task and obstacle primitive to each robot. (\textbf{D}) A simplified training cell with a reduced number of obstacles and tasks. Each task is specified and visualized as a coordinate frame that the end-effector of the robots must match, where the blue/Z axis is inline with the radial axis of the end-effector.}
\label{fig:scene_graph}
\end{figure}

However, naive applications of deep RL to multi-arm task and motion planning would also fail to scale beyond trivial scenes, as the scene complexity still scales combinatorially and this leads to exponential scaling in the required size of the underlying deep neural network representing the agent's policy. The main contribution of this work is in the use of graphs to represent the state, and the use of graph neural networks (GNN) for RL policy and state-action value estimation. GNNs use weight sharing to allow for scaling the graph size without scaling the model complexity, similarly to how convolutional neural networks (CNN)\cite{lecun1989handwritten} have fixed kernel sizes independent of image sizes, and inference time complexity is quadratic in the number of robots - refer to the Discussion section for more details. The state graphs are constructed such that graph nodes represent robots, obstacles and tasks, and graph edges connect each task and each obstacle to each robot, and each robot to each other, supporting coordination amongst the robots and allowing each node to receive sufficient information about the environment to plan a collision-free path to tasks (Figure \ref{fig:scene_graph}).

With this formulation, RoboBallet directly and simultaneously controls multiple (here up to eight) 7-Degree of Freedom (DoF) robots, coordinating an up to 56-dimensional configuration space, and handling up to 40 shared tasks, with completely unconstrained robot allocation and task scheduling. The model decides at each 100ms time-step the target joint velocities of all joints in all the robots, and our environment setup simply applies velocity and acceleration clamping, stops robots from colliding, and integrates the resulting velocities to generate joint configuration waypoints.

RoboBallet adopts an approach that prioritizes scalability and trades off theoretical guarantees, so as to enable real-world deployment on a much larger scale of problems (test with up to eight robots in dense environments) while demonstrating empirically high quality solutions. RoboBallet opens up new and exciting possibilities for many industrial applications, such as workcell layout optimization, fast re-planning in case one robot fails, and handling run-to-run variations in work piece pose using image-based pose estimation to capture the work piece pose for the planner, as well as environment changes using 3D environment reconstruction with cameras or LiDARs. RoboBallet is an enabling technology that opens the door to these important capabilities that have thus far not been possible due to the high cost and latency of manual task and motion planning in the multi-robot manufacturing setting.

\section*{RESULTS}

To demonstrate the capability of RoboBallet, we trained the model over several days on randomly-generated workcells of different sizes in simulation, and validated the performance of trained models in a simulated workcell with eight robot arms, a work piece constructed out of aluminum struts, and 40 tasks (Movie \ref{fig:train_eval_setup}). To demonstrate the feasibility of real-world execution of our generated trajectories, a model with four robot arms was trained using exactly the same framework, and we validated the generated trajectories on a real workcell with minimal post-processing.

We observed that our models generated high-quality trajectories in seconds and can effectively use many robots to shorten execution time even on the largest settings tested (eight robots, 40 tasks, 30 obstacle primitives), on a single GPU, in both a real and a simulated multi-arm workcell.

To showcase one of the most important capabilities enabled by a fast automated planner, we used RoboBallet, combined with a black box optimizer, to perform workcell layout optimization (optimizing placement of robots given fixed sets of tasks and obstacles). We saw that this optimization setup improving task execution times by up to $33\%$ compared to non-optimized robot placements in our test cell, on our fixed test set of tasks.

The training environments had either four or eight Franka Panda 7-DoF robots: four mounted on a table, and an additional four mounted on the ceiling mirroring the four on the table in the case of eight robots (Movie \ref{fig:train_eval_setup}). At the beginning of each episode, we first randomized the location of the robots along virtual 1.6m-long rails parallel to the side of the table, only rejecting configurations that are in collision at the start. Then we sampled and placed 30 cuboid obstacles with randomly generated positions and orientations, only rejecting obstacles that would be in collision with the robots in their start configurations. Next, we uniformly sampled tasks on the surfaces of the obstacles with a small offset away from the surfaces, where each task is a position and an orientation which a robot end-effector must reach. We then tried to generate IK solutions from each robot to the task pose, and rejected the task if it does not have at least one collision-free IK solution from at least one robot. Information about the IK solutions and reachability were not given to the model.

In the motion planning literature, the execution time (the durations of trajectories) is the primary metric to be optimized for. However, as no published approach exists that can scale to our problem sizes in the multi-arm task and motion planning setting with realistic execution time, we provide a comparison against a baseline method for trajectory optimality in a simplified setup. This is given in the \textbf{Optimality of generated trajectories} sub-section. The rest of the \textbf{RESULTS} section reports on the scalability and generalizability of the algorithm on the full problem. For a qualitative evaluation, please refer to Movie \ref{fig:real_robot_snapshots} as well as Supplementary Videos S1-S4 showing generated trajectories.

\subsection*{Optimality of generated trajectories}

In both real-world applications and academic benchmarks, trajectories are most commonly evaluated by their time cost (duration). In this section, we tested our approach on a set of smaller and simplified problems where we were able to generate near-optimal trajectory solutions through exhaustive search and pre-assignments of tasks to robots. The pre-assignment is necessary as exhaustively testing all possible allocations in addition to schedules is not computationally feasible for all but the smallest problems. 

Specifically, we generated 10 sets of 20 tasks, where each set contains five tasks for each of four robots. We filtered the sampled tasks to ensure that they can each only be reached by one robot. To generate solutions that are as close as possible to optimal to compare against, we needed to tackle the same four challenges that our model needs to tackle simultaneously - task allocation, task scheduling, IK solution selection, and motion planning. Optimal task allocation is ensured by problem design (only one allocation is possible). Optimal task scheduling is done by exhaustively testing all possible task schedules. In the case of 7-DoF robot arms, each 6-DoF task pose has up to an infinite number of IK solutions when poses are within the feasible reachability workspace, and choosing good IK solutions in the context of each trajectory is also important in minimizing total trajectory cost. We present multiple minimum costs based on different numbers of IK solutions sampled (and exhaustively tested) to compare against. Finally, as a computationally-feasible optimal motion planning algorithm for 7-DoF does not exist, we used RRT-Connect\cite{kuffner2000rrt} with shortcutting, which is a widely-used motion planning algorithm known for generating low cost trajectories in practice. We first planned for each robot separately, then, for each task set, we took the cost of the longest single-robot sub-trajectory as the cost of the multi-robot trajectory.

\begin{figure}[H]
\centering
\includegraphics[width=15cm]{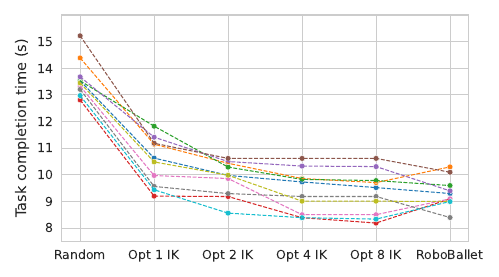}
\caption{\textbf{Comparison against RRT-Connect + Exhaustive Task Scheduling Baseline in a simplified setting.}\\
Each colour represents one task set with 20 tasks (5 per robot). All columns except 'RoboBallet' uses RRT-Connect and shortcutting for motion planning. The 'Random' column averages over all possible task schedules. The 'Opt [N] IK' columns uses N IK solutions per task, and the optimal task schedule and IK solution selection by exhaustive search.}
\label{fig:baseline}
\end{figure}

In Figure \ref{fig:baseline}, we first showed that with a random task schedule for each robot, the trajectory costs are much higher than with optimal task scheduling, demonstrating the importance of good task scheduling even in this simplified setting. Then we showed that by increasing the number of IK solutions sampled (and selecting the best IK solution for each task in each trajectory), the trajectory costs also decreased substantially, demonstrating the importance of choosing good IK solutions. Finally, we showed that RoboBallet generated trajectories with costs that are competitive with the baseline approach at 8 IK samples per task. These results showed that although an RL-based approach is not able to offer the probabilistic-completeness that sampling-based methods do, the costs of trajectories generated in our problem settings are comparable with our baseline results in practice. The baseline also uses orders of magnitude more compute and relies on several simplifying assumptions, unlike RoboBallet.

There are two notable differences between the baseline and RoboBallet setups that need to be taken into account while interpreting these results. Firstly, the RRT-Connect algorithm builds a tree from both the start and the finish poses, generating a valid path that puts the end effector at exactly the goal pose, whereas our approach uses a tolerance of $2.5$ cm and $15^{\circ}$. In a real-world application, our approach would integrate a linear driver to achieve final docking. This would be implemented as part of the environment, where the RL agent only receives a reward if the final docking is successful, ensuring that the agent chooses solution poses where the final docking is feasible. This is excluded from the current study for simplicity, and its inclusion would result in a marginal increase in trajectory costs for RoboBallet. Secondly, the RRT-Connect results are generated with each robot in isolation (other robots removed), whereas RoboBallet moves all robots at the same time. This means that the RRT-Connect results here ignore potential collisions between robots during task execution. If collisions did occur in practice when all robot trajectories are executed simultaneously, this would require adjusting task sequences to create a collision-free multi-robot trajectory.

\subsubsection*{Training time scaling in the number of robots and tasks}
\begin{figure}[H]
\centering
\includegraphics[width=7.3in]{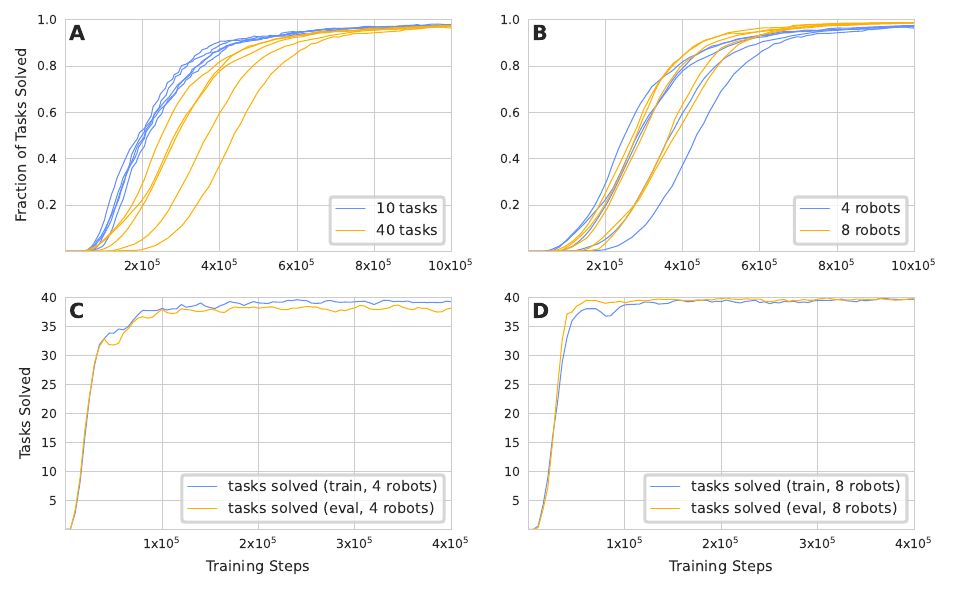}
\caption{\textbf{Training Curves for Scalability and Generalizability.}\\
(\textbf{A}) \textbf{Training curves for 10 versus 40 tasks, four robots}. (\textbf{B}) \textbf{Training curves for four versus eight robots, 40 tasks}. Note that the number of training steps required to reach convergence does not scale exponentially. However, in terms of asymptotic time complexity, the amount of computation required is linear in the number of tasks, and quadratic in the number of robots. (\textbf{C}) \textbf{Results on randomly generated training cells versus hand-designed evaluation cell over training, four robots}. (\textbf{D}) \textbf{Results with eight robots}.}
\label{fig:curves}
\end{figure}

As discussed in the Introduction, classical approaches scale exponentially in the number of robots and number of tasks, and this prevents them from being applicable to large settings. In Figure \ref{fig:curves}, we demonstrated that our approach scales much more favorably at training time. Quadrupling the number of tasks only required marginally more training steps to converge, despite the ${4 \times}$ difference in the number of tasks, and therefore in the difficulty of the scheduling problem. Note that this does not correspond to equal amounts of computation. The number of edges in our scene graph is linear in the number of tasks, and therefore, each training step takes up to ${4 \times}$ more compute in the 40 tasks case compared to the 10 tasks case.

The second important aspect of computational scalability pertains to the number of robots. Similarly to the scaling with number of tasks, we saw that the training curves converged at a similar number of training steps. The theoretical asymptotic time complexity is $O({N_{robots}}^2)$, as the robot nodes are fully connected with each other for coordination. In practice, quadratic scaling here is not a problem as there is an inherent limit in the density of placement of robots, beyond which there would be no benefit in execution time as it becomes more and more difficult for the robots to avoid one another. For very large workcells with more than a dozen robots, the robots would be positioned such that they would not all have overlapping work volumes. If two robots do not have overlapping work volumes they do not need to coordinate with each other, as they cannot complete the same tasks nor collide with each other. Therefore, in the asymptotic case, assuming we have a limiting density in placement of robots, and add more robots by expanding the workcell without connecting robots that do not have overlapping work volumes, the algorithm's asymptotic time complexity falls back to $O(N_{robots})$.

In practice, due to the performance characteristics of GPUs, the run-time differences are even smaller than suggested by the asymptotic complexity. In our case, each training step took 33 milliseconds for the four robot and 10 tasks case, 52 milliseconds for the four robots and 40 tasks case, and 80 milliseconds for the eight robots and 40 tasks case, with a batch size of 128 state transitions per training step. Figure \ref{fig:curves} shows the training curves with three different configurations - four robots with 10 versus 40 tasks, and 40 tasks with four versus eight robots.

\subsubsection*{Inference time scaling}
At inference time, even the largest setup with eight robots and 40 tasks only took about 0.3 milliseconds per planning step on an NVIDIA A100 GPU, allowing for greater than 300x real-time planning at 10 Hz time-steps. On a single Intel Cascade Lake CPU core, each step takes about 30 milliseconds, which is still 3x faster than real-time with 10 Hz time-steps. The planning process consists of one inference and one collision check for the entire scene at each time-step.

\subsubsection*{Efficient use of multiple robots}

\begin{figure}[H]
\centering
\includegraphics[width=12cm]{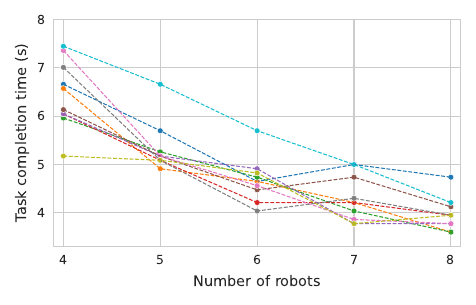}
\caption{\textbf{Task completion time scaling with the number of robots (optimized placements).}\\
Shortest trajectory execution times achieved with the 10 evaluation configurations using optimized placements of four to eight robots. Each color represents one evaluation configuration.}
\label{fig:robot_scaling}
\end{figure}

Another important aspect of scalability is the reduction in task completion time as the number of robots increase. The ability to control more robots is only meaningful if the controller is able to solve tasks faster when given more robots to use. For this experiment, we first randomly sampled 10 sets of robot placements, each with four robots - two on either side of the table, and 2 on either rail on the ceiling. For each set of placements, we sampled a set of 20 tasks that can be solved using those four robots in those randomly sampled placements. These sets of tasks made up the evaluation configurations, each of which is a task set that is guaranteed to be solvable using four robots. Then, we performed experiments to find the shortest trajectory lengths we can generate for each evaluation configuration using four, five, six, seven, and eight robots. For each number of robots and evaluation task set, we performed placement optimization (see \textbf{Workcell layout optimization}) to find the best placements for the given number of robots and set of tasks, and recorded the minimum trajectory length achieved. Figure \ref{fig:robot_scaling} shows the best execution time achieved using four to eight robots with optimized placements for each evaluation configuration. The plot shows strong execution time scaling from four to eight robots, with the average execution time reducing from approximately 7.5s using four robots, to 4.5s using eight robots (a reduction of $60\%$). This is approximately optimal scaling, given that in the four robots case, with uniform task distribution, each robot needed to perform five tasks, whereas in the eight robots case, four of the robots needed to perform two tasks, and the other four needed to perform three tasks.

\subsection*{Generalizability to hand-designed environments}

Our models were trained with randomized robot placements, randomized obstacles, and randomized tasks to enable maximum flexibility at evaluation time. Movie \ref{fig:train_eval_setup} shows the evaluation setup, where the obstacle is a hand-designed shape made out of aluminium struts, robot placements are fixed to the locations in our real test workcell, and the tasks are randomly sampled on the faces of the obstacle. We generated 10 sets of 40 tasks for evaluation, which are not used in training. Panels C and D in Figure \ref{fig:curves} show performance on these evaluation configurations compared to the training environment, over the training process. Our results show that the model, trained on randomly generated environments, can generalize very well to an unseen hand-designed evaluation cell.

\subsection*{Workcell layout optimization}

A fast task and motion planner enables use cases that are not currently possible with manual planning. One such use case is workcell layout optimization. In this experiment, we trained the model on fully randomized environments as described above, and used the Google Vizier\cite{vizier} black-box optimizer with a modified Gaussian process bandit algorithm\cite{song2024vizier} to find the best placement of robots to achieve the shortest execution time on each of the 10 sets of fixed evaluation tasks. Figure \ref{fig:layout_optimization} shows the execution time improvements from the optimized placements compared to the common default placements with one robot at each corner of the table, as well as the breakdown of time spent on each task by each robot, both with motion plans generated using RoboBallet. We see that there is universal improvement in trajectory execution time over all evaluation configurations, and that the improvement comes primarily from the more even redistribution of tasks to robots, and secondarily from reduced motion time to reach the task poses. In setups where the task completion time is long relative to the motion time, and therefore the even distribution of tasks is more crucial, the benefit of placement optimization would be even more pronounced. We chose to use a dwell time of 0.5s (simulating task completion time) at each task here to show the result on a more balanced setup where both motion time and task completion time contribute substantially to the total solution length.

\section*{DISCUSSION}

Our results demonstrate that graph neural networks combined with reinforcement learning enable efficient solutions to the joint problem of task allocation, scheduling, and motion planning. Although the conventional sampling-based approaches offer probabilistic completeness -- a valuable property in certain contexts -- our approach is able to scale to much larger real-world problem sizes, delivering high quality solutions with minimal computation requirements, especially at evaluation time. This trade-off between theoretical completeness and scalability is critical for practical applications in large-scale, dynamic environments.

The high scalability of our method comes from the fact that in terms of model complexity of our learned functions, the network only has to learn the interactions between different types of nodes (tasks to robots, obstacles to robots, and robots to robots) once. This results in $O(1)$ scaling in model complexity (model size) with respect to the number of robots, number of tasks, and complexity of obstacles. This means that theoretically, the amount of training data needed does not scale with the size of the environment, and we observed that to be broadly true in our experiments. In terms of asymptotic time complexity, computation is dominated by the edge processing, which is proportional to the number of edges. We have ${N_{robot} \times N_{robot}}$ robot-to-robot edges, ${N_{robot} \times N_{task}}$ task-to-robot edges, and ${N_{robot} \times N_{obstacle}}$ robot-to-obstacle edges, for a total time complexity of $O({N_{robot}}^2 + N_{robot}N_{task} + N_{robot}N_{obstacle})$. This complexity is linear in both number of tasks and number of obstacles, and quadratic in the number of robots, which is manageable due to the inherent limit in how many robots can be usefully placed in the same space.

A fast and automated task and motion planner enables many new use cases. As an example that we explored in this work, we used the planner to design the layout of new robot workcells by running a black-box optimizer on top of the planner, to find the best robot placement for each set of tasks. The cost of robots and floor space is one of the primary limiting factors in many manufacturing setups, and our results show that we are able to substantially reduce the execution time of each of our evaluation configurations by performing layout optimization. With our reinforcement learning setup, the reward function can easily be modified to incorporate different optimization objectives with different importance. As an example, for our demonstrations on real robots, we trained the model with an additional reward term for returning to the start position at the end of the episode using Euclidean (L2) norm in configuration space from the start configuration, only activated once all tasks are solved. We also used a weak penalty term based on squared joint space acceleration, to discourage superfluous movements and unnecessarily extreme accelerations. Robotic applications often have unique requirements, and with an RL formulation, we only need to design a cost function to reflect them, which is often much easier than designing behaviours, making RL-based systems more easily adaptable to different requirements compared to both hand-coded systems, and supervised-learning based systems learning from either an algorithmic planner or human demonstrations.

As another example use case, in situations where assembly line halts due to robot failures are not acceptable, the only strategy currently available in the industry is to create fail safe motion plans with each robot removed, which is extremely costly in the manual labour required to design these plans, and not scalable to more than one robot failure per cell. With our approach, the trained model can either generate the backup trajectories ahead of time, or as the failures happen, with minimal downtime. In case the remaining robots cannot continue to cover all the tasks, but at least some of the robots can be moved (e.g. along linear rails), the model can also be used, with layout optimization, to quickly determine the optimal new placements for the remaining robots.

In terms of real world execution of the generated trajectories, although the models already respect joint velocity and acceleration constraints, it would be prudent to perform post-processing on the generated trajectory for smoothing and for compliance with more complex constraints such as jerk and torque. Movie \ref{fig:real_robot_snapshots} shows trajectories generated by the model post-processed by 5-th order spline interpolation to 1 kHz running on Franka Panda robots. In this case the network has also been trained to respect a 1.0 cm collision margin for additional safety.

High density robotic task and motion planning for reaching is a problem that expert humans can manually solve at great time investment and man-power costs, using human experience and intuition to navigate the huge combinatorial search space. But this problem has proven to be very challenging for automated algorithmic solutions. Machine learning provides methods to imitate human intuition that is hard to express as hand-coded heuristics. However, a naive machine learning-based solution would suffer the same fate of combinatorially increasing cost, if the model architecture does not take into account the relational inductive biases that can be exploited to reduce the required learning efforts, and improve generalization. Our combination of a structured state representation and a model learning approach that exploits the structure mirrors how expert humans solve this problem using their intuitive understanding of the construction of the problem.

The physical world is inherently structured, and one of the most intuitive computational representations of those structures is as graphs. In our case, we chose to represent the scene as a graph made out of robots, tasks, and obstacles nodes, exploiting the fact that each type of interaction (e.g. driving a robot to a task, or avoiding an obstacle) only needs to be learned once, regardless of the complexity of the scene. This gives us better data efficiency, as well as fewer parameters to learn, compared to naive network architectures that do not exploit these inductive biases. The fact that we can learn from every interaction of the same type (e.g. robot to task) to train one fixed size function turns the challenge of combinatorial complexity into an advantage, and gives us the ability to efficiently train small and high quality models that do not over-fit to the training data.

\subsection*{Limitations and future work}

In practical applications, the key challenge we identified in task and motion planning for dense robotics is to achieve scalability, and it is the main factor preventing existing solutions from being applicable to real-world applications. Our approach is demonstrated on a general reaching setting and a formulation where any robot can complete any task in any order within a shared workspace. To specifically focus on the scalability challenge, the configuration in this paper represents the largest search space, allowing us to tackle the fundamental scalability issues before introducing additional constraints. In practical/industrial applications, there are often constraints such as inter-task dependencies and heterogeneous robot capabilities which are not considered in this work. Our approach could be extended to handle these constraints, for example, by disabling edges between incompatible robots and tasks, or adding edges between tasks to signal dependencies. The resulting problem is theoretically simpler to solve than the fully unconstrained problem, but this is to be demonstrated in future work. Similarly, on the motion planning level, we chose the abstract setting where the tasks are fully constrained end-effector poses, which makes motion planning the most constrained and difficult. In many settings such as spot-welding or painting, the task poses are only semi-constrained, e.g., some rotational constraints of the end-effector in spot welding are soft-constraints. We believe this can be accomplished by augmentation of the reward function, and that is also left to be demonstrated in future work.

The favorable scaling properties demonstrated in this paper are sufficient to enable many applications and industrial use cases, but if scalability to even larger settings is required, it would be computationally more efficient to apply some heuristics to prune graph connectivity. For example, a pre-processing step can determine reachability between all robot and task pairs, and edges are only added where a robot can reach a task. This takes $O(N_{robot}N_{task})$ time, and would not increase the time complexity of the overall algorithm. Similarly, where a robot's work volume does not overlap with an obstacle, the robot can never collide with the obstacle, and the pair would not need a connecting edge. We did not implement these simple optimizations as we wanted to explore the limit of machine learning and keeping the algorithm as simple and generally applicable as possible, but in a real-world application it would be prudent to include them. A graph attention mechanism \cite{velickovic2017graph} may also improve training in very large cells to automatically reduce the relevant pairs of entities the network has to consider.

For simplicity we chose a minimal representation of obstacles -- cuboids. Although this works well for our evaluation environments, it may not be suitable for environments where most obstacles cannot be decomposed into a reasonable number of cuboid shapes. In that case, and especially if the robots must get very close to the obstacles (e.g., for spot-welding), it may be beneficial to add more primitive types, and pick the best primitive type for each decomposed obstacle component. This approach would require an adjustment in training to randomly place obstacles of all primitive types. Another possibility is to support arbitrary polyhedrons by representing them as graphs themselves, and using another graph neural network to summarise their properties into fixed length embeddings to be used in the main neural networks. The embedding and policy/value networks can be trained end-to-end to ensure that the embeddings capture useful information for motion planning.

To clearly illustrate the core capabilities of our approach, we focus on reaching tasks. However, the algorithm can be readily extended to more complex operations, such as pick-and-place tasks. One possible formulation is to factor pick-and-place tasks into separate picking and placing task poses, adding and removing task dependencies and compatibility as required using edge features in the state graph, and have the simulator attach and detach objects from end effectors where necessary. In this formulation, a separate controller or a grasp pose predictor is required to generate feasible pose targets for picking up arbitrary objects, both at training time and at inference time. Extension to manipulation tasks presents further challenges, including object collisions, robot-environment constraints, and occlusion handling in multi-robot coordination, requiring future work on both algorithmic development and experimental validation.

Going beyond existing capabilities into potential future applications, although our method is more than fast enough at inference time for online planning, online planning comes with other challenges such as perception and robot trajectory tracking accuracy without post-processing. Solving these issues, together with our fast planner, would enable industrial capabilities that are not currently available, and would be a valuable extension to this work.

As an extension to our workcell layout optimization example, using RoboBallet combined with a black-box optimizer as the outer loop, a designer can be use it to quickly find optimal number, types, and positioning of robots in a workcell. For example, they may need to make a decision between using a 6-DoF robot mounted on a linear rail, versus a 7-DoF robot, or maybe a heterogeneous team of robots. In this paper, we have presented a simplified version where only the placements are optimized. If the choice of robots is also to be optimized, the model would be trained with robot morphology randomization at training time, in addition to the other axes of randomization, so that the resulting model would be compatible with multiple robot morphologies.

Finally, bi-manual manipulation is the most natural way for humans to interact with the environment, and enables capabilities that are out of reach of single manipulators. One of the main challenges in multi-arm manipulation is the combinatorial complexity in performing motion planning in such a high dimensional action space. Our approach shows that this scalability can be achieved using a structured network architecture, which can provide more insights and pave a way to develop future technologies in bi-manual and multi-manual task settings, extending the nature of multi-arm coordination from collision avoidance, which is the focus of our work, to cooperative task completion in many other domains of robot manipulation.

In this work we presented a method for multi-robot task and motion planning (demonstrated on but not fundamentally restricted to reaching) that is able to perform automated planning on industrial-scale workcells, a task that is currently performed manually at great cost and great latency. We showed that a fast planner like ours not only allows for accelerating this process, but also makes new capabilities such as workcell optimization and online planning feasible.

\section*{MATERIALS AND METHODS}

Our goal is to train a neural network-based controller that, given the states of all robots, tasks, and obstacle primitives, will generate a set of joint velocities for each robot such that, when followed over time, produces the most efficient trajectory to solve all the tasks. To that end, we designed a reinforcement learning formulation, where the observations are represented as graphs in which nodes represent robots, tasks, and obstacle primitives, and edges represent the relationships between them (e.g. relative-pose). The trained policy network is a graph neural network\cite{battaglia2018relational} that takes the observation graphs as input, and produces commanded joint velocities for all robots at each time-step. This in effect allows the agent to reason relationally and combinatorially, while only taking in raw state as inputs. The joint velocities are integrated to produce next robot configurations, except where the commanded velocities would cause a collision, in which case the joint velocities for the robot is zeroed. The agent is rewarded for solving tasks, and avoiding collisions. We use a slightly modified asynchronous version of the TD3\cite{fujimoto2018addressing} RL algorithm to train the policy network as well as two auxiliary discounted state-action return prediction Q networks, also known as critics. The overall training and evaluation setup is shown in Figure \ref{fig:scene_graph}.

\subsection*{Reinforcement learning setup}

\subsubsection*{Environment}

The environment is implemented using a kinematic simulator. Each action applied advances the simulation by 0.1 seconds, and each action consists of desired joint velocities for all joints of all robots. We apply acceleration and velocity limiting based on robot kinematic limits. A task is considered solved if the end effector of a robot moves to within an Euclidean distance of $2.5$ cm and angular distance of $15^{\circ}$ of the task pose. Although this may seem coarse, there is a trade-off between tolerance and learning speed, and in practice 2.5cm is close enough to generate good plans which can be post-processed to arbitrary precision. When a task is solved, the solving robot is frozen for 0.5 seconds dwell time to simulate performing some work at the site, and also to encourage even task distribution between robot arms. If the joint velocities generated for any of the robots would cause it to go into collision with itself, another robot, or the environment, the joint velocities for the robot in question is zeroed, and thus the robot stands still for that time-step. Therefore, the environment does not allow for collisions. This is a desirable property for transfer to real world setups where collisions must be avoided at all costs. We use a collision penalty to discourage the policy from generating joint velocities that would cause collisions (see the section on below on rewards for more details).

\subsection*{Observation}

We represent environment states as graphs, with robots, tasks, and obstacle primitives as nodes. The primary benefit of using this structured representation rather than simply flattening the state information into a vector is that it provides a way to incorporate relational information between entities in a reusable way, and allows the network to reason about them relationally.  For example, the model only has to learn to reason about relative-pose between a robot and a task once, and it automatically transfers via the graph to all combinations of robots and tasks. By contrast, conventional architectures would be forced to discover these relationships for every combination of entities, leading to exponential compute and memory scaling, as well as poor data-efficiency.

In GNNs edges can be directional -- we use bi-directional edges between robot nodes to allow for collision avoidance and coordination, whereas uni-directional edges from the task nodes and obstacle nodes to robot nodes give the necessary information to robot nodes for motion planning. The directionality of edges determine how information flows during the graph inference step, and our choice was designed to propagate information to the robot nodes to allow reading out the actions for each robot.

\subsubsection*{Obstacle representation}

Our goal is to train a generalised planner that can plan in environments it has not seen during training, and supports obstacles with arbitrary geometries. That requires breaking obstacles down to primitives, where each primitive can be added to the graph as a node. During training, we generate cuboid obstacles of random sizes, positions, and orientations to add to the graphs as obstacle nodes, filtered only to exclude collisions between them and the robots in their initial poses.

\begin{figure}[H]
\includegraphics[width=7.3in]{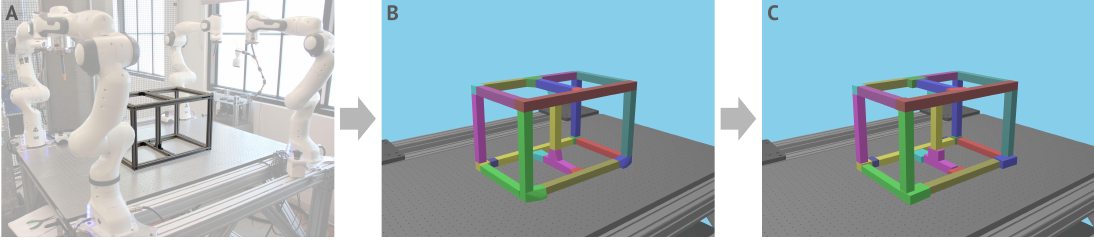}
\caption{\textbf{Obstacle mesh decomposition pipeline.}\\
(\textbf{A}) Starting from the CAD model of the obstacle. (\textbf{B}) Result of decomposition into convex polyhedrons using V-HACD. (\textbf{C}) Oriented bounding boxes computed from the convex polyhedrons using \cite{chang2011fast}.}
\label{fig:decomposition}
\end{figure}

At evaluation time with externally-defined meshes as obstacles, the obstacle meshes are decomposed into approximate convex hulls, and the oriented bounding boxes of those convex hulls are used as obstacle nodes (see Figure \ref{fig:decomposition}). The approximate convex decomposition is performed using the V-HACD library\cite{vhacd}, whereas the oriented bounding box search uses our implementation of the algorithm described in \cite{chang2011fast}, which consists of a combination of genetic and Nelder-Mead\cite{olsson1975nelder} algorithms to efficiently find the best rotation for minimum volume bounding boxes through unconstrained optimization on the 3D rotation group $SO(3)$. Since each cuboid has four equivalent representations that can be created by exchanging axes and making the corresponding change in pose, we randomly select between the representations at each time-step as a simple augmentation.

\subsubsection*{Task sampling}

At training time, the tasks (desired end-effector poses), are sampled uniformly in position from the surfaces of the obstacle cuboids, and in orientation randomly such that the desired orientation corresponding to the end effector direction is maximally $22.5^{\circ}$ from the normal direction of the face (that is, to complete the task, the end effector will be pointing approximately into the obstacle). At evaluation time, the tasks are generated in the same way on the externally-defined obstacle geometry. We check that each task pose has at least one collision-free IK solution from one robot. For efficiency we are not verifying that there is a collision-free path from the start position to the task pose, only that a collision-free IK solution exists. Therefore, some of the tasks may not have a valid path solution. The IK solutions are not stored and no information from this task filtering process is used in the training process.

\subsubsection*{Node features}
The environment observations are in the form of graphs, where robot and task nodes have different features, whereas obstacle nodes have no node features. The node features for different types of nodes are then concatenated in a block-diagonal fashion to create the combined node features for the scene graph.

For each robot, the robot node consists of the \textbf{joint configuration} for each of the 7 DoF of the robot, \textbf{joint velocity} for each DoF, and \textbf{dwell time remaining} which is non-zero if the robot has reached a task pose, and is executing the task (we freeze the robot in the task pose for 0.5 seconds dwell time to simulate executing some task at the site).

Each task node consists of a single feature: \textbf{task status} that is recording whether a task has been solved.

\subsubsection*{Edge features}

Edge features encode relative information between each pair of nodes where an edge exists. Similarly to the node features, we have different edge features for each edge type. In a graph net, edges are directional and represent direction of information flow, and we use ``sender" and ``receiver" to denote the vertices of an edge relative to the direction of information flow. All poses are encoded using a 3D translation and a 6D representation of the rotation\cite{zhou2019continuity}.

The robot $\rightarrow$ robot edges encode the sending robot base pose relative to the receiving robot tip pose. The task $\rightarrow$ robot edges similarly encode the task end effector pose relative to the receiving robot tip pose. Finally the obstacle $\rightarrow$ robot edges have the following features: \textbf{centroid translation of the obstacle primitive} relative to the receiving robot tip pose, and \textbf{rotation matrix} of the obstacle primitive relative to the receiving robot tip pose, with each basis vector scaled by the corresponding span of the primitive in that direction.

\subsubsection*{Global features}
The global features of the graph consist of the following: \textbf{episode time} normalized to timeout, the episode time at which we terminate the episode if the tasks have not been completed, and the \textbf{current score} as defined in the Reward section.

\subsubsection*{Reward}

To calculate the reward for a step we first calculate a score for the state, which is simply the fraction of tasks that have already been solved (Eq. \ref{eq:tasks_score}),

\begin{equation}\label{eq:tasks_score}
S(s) = \frac{N_{done}(s)}{N_{total}(s)} .
\end{equation}

The reward consists of two components. The first component is the difference in score between the new state and the previous state (Eq. \ref{eq:score_reward}),

\begin{equation}\label{eq:score_reward}
R_{score}(s, a) = S(s + a) - S(s) .
\end{equation}

The second component is a penalty for actions that would have put the robots in collision (if it was not stopped by the environment). This is the number of robots that would be put in collision, scaled by a coefficient $C_{col}$ (Eq. \ref{eq:collision_penalty}). Although in theory, we do not need to add an additional penalty for collisions, as the increase in execution time, combined with the discounting, would encourage the agent not to command joint velocities that would cause collisions, which are not optimal. However, in practice, adding an explicit penalty for collisions is effective to both reduce collisions and improve solution execution time:

\begin{equation}\label{eq:collision_penalty}
R_{collision}(s, a) = C_{col} N_{robots\_colliding}(s, a) .
\end{equation}

The total reward is the sum of the two components (Eq. \ref{eq:total_reward}),

\begin{equation}\label{eq:total_reward}
R(s, a) = R_{score}(s, a) + R_{collision}(s, a) .
\end{equation}

\subsubsection*{Hindsight Experience Replay}

As the reward as specified above is sparse, it is unlikely that an RL agent early on during the training process would be able to find good rewards through random exploration. The conventional solution to this problem is to engineer a shaped reward - by giving partial credit when the agent moves robots closer to completing tasks. However, designing a good shaped reward that encourages fast and reliable learning is not trivial, and is also specific to the type of task being performed. Hindsight experience replay (HER), as proposed in \cite{andrychowicz2017hindsight}, provides another solution to the sparse reward problem. With HER, instead of providing partial credits to aid learning when the agent is not yet reaching tasks, with each failed episode, we also generate a corresponding imaginary episode where we move a randomly selected subset of tasks to randomly selected points of the robot trajectories within the episode, therefore providing positive training episodes that in essence shows the learner that it would have been a successful episode, if the tasks were actually along the robot trajectories the agent generated. This provides rich training signals with a healthy mix of successes and failures, for efficient learning. Please refer to \cite{andrychowicz2017hindsight} for more details on the general concept of HER.

In our experiments we found that HER performs approximately as well as even a well-tuned shaped reward, so we relied entirely on HER for the experiments above for simplicity.

\subsubsection*{Modified Twin-Delayed Deep Deterministic Policy Gradient (TD3)}

TD3\cite{fujimoto2018addressing} is an RL algorithm based on the deterministic policy gradient algorithm\cite{silver2014deterministic} in an actor-critic framework, where two networks are trained simultaneously - a policy network $\pi(s)$ that predicts the best action to take from a state, and a critic network $Q(s, a)$ that predicts the discounted total future return (Q-value) given a state and an action taken in that state. The policy network is trained to produce actions that result in a high Q value,

\begin{equation}\label{eq:policy_update}
\pi(s) \Rightarrow \mathrm{argmax}_{a} Q(s, a),
\end{equation}
and the critic network is trained to predict the sum of the immediate reward from applying action $a$ in state $s$ and the discounted Q-value from the resulting state $s'$, with the best action in that state. Since we do not know the best action in state $s'$, we use the output of the policy network $\pi(s)$ as an approximation,

\begin{equation}
Q(s, a) \Rightarrow r(s, a) + \lambda Q(s', \pi(s')),
\end{equation}
where $(s, a, r, s')$ is a transition from applying action $a$ to state $s$, arriving at state $s'$ with reward $r$, and $\lambda$ is the discount factor.

Deterministic policy gradient algorithms are a newer class of RL algorithms where the policy network predicts an action directly, as opposed to a distribution over actions that would be predicted by policy networks in earlier stochastic policy algorithms. They are more efficient to train, but are prone to get stuck in local minima. TD3 is the current state of the art of a lineage of deterministic policy gradient algorithms that each added some tweaks to stabilise training. We refer the reader to \cite{fujimoto2018addressing} for a thorough description of the algorithm.

In our implementation of TD3, we removed the delayed target network updates for simplicity, as similarly to what the authors reported, we found the stability contribution of the delayed updates to be minimal, and the benefit is mainly a computational one. However, as our network architecture is much more complex, the target network updates do not constitute a notable part of the compute requirement.

We also decoupled critic and policy Polyak averaging\cite{polyak1990new} constants so they can be tuned independently. On the actor side, we introduced exponential decay in exploration variance, to encourage strong exploration in the beginning, and more accurate task solutions later on in the training process. Similarly we introduced two exponential decays - $\lambda_{\eta_{\pi}}$ and $\lambda_{\eta_{Q}}$ for the policy and Q-function learning rates, respectively.

\subsection*{Network architecture}

The system consists of three learned functions - two Q-value estimators, and a policy function. All three functions use a graph neural network $\textrm{GNN}$ at their core, with the same architecture, but initialised and trained independently without weight sharing.

\subsubsection*{Graph neural network core}

The core GNN takes as input graph $G$ with nodes $N$ with features matrix $G_n$, edges $E$ with features matrix $G_e$, and graph-global feature vector $G_g$, and produces an output graph $G'$ similarly with node, edge, and global features $G_n'$, $G_e'$, and $G_g'$. The GNN contains three learned update functions - a node update function $U_n$, an edge update function $U_e$, and a global update function $U_g$. In all of the following equations, $\oplus$ denotes concatenation.

The GNN first gathers, for each edge $e \in E$, the sending and receiving node features $G_{n}[n_s]$ and $G_{n}[n_r]$ (where $n_s$ is the sending node, and $n_r$ the receiving node), edge features $G_e[n_s \rightarrow n_r]$, and global features $G_g$, and concatenates them for input into the edge update function $U_{e}$, which generates new edge features $G_e'$ for the output graph (Eq \ref{eq:edge_update}): 

\begin{equation}\label{eq:edge_update}
G_e' = U_{e}(G_{n}[n_s] \oplus G_{n}[n_r] \oplus G_e[n_s \rightarrow n_r] \oplus G_g) \quad \forall e (n_s \rightarrow n_r) \in E .
\end{equation}

Then, for each node $n \in N$, all the edge output features for edges coming into the node are element-wise aggregated by summation to generate an aggregated view of information arriving into the node. This is concatenated with the node input features and the global input features, and passed as input into the node update function $U_{n}$, which generates node output features $G_n'$ (Eq \ref{eq:node_update}): 

\begin{equation}\label{eq:node_update}
G_n' = U_{n}((\sum_{e \in E(\ast \rightarrow n)} G_e'[e]) \oplus G_n[n] \oplus G_g) \quad \forall n \in N .
\end{equation}

Finally, the global update function $U_{g}$ takes as input the aggregated node embeddings and edge embeddings, as well as the input global features $G_g$, and generates new global embeddings $G_g'$ (Eq \ref{eq:global_update}):

\begin{equation}\label{eq:global_update}
G_g' = U_{g}(\sum_{n \in N} G_n'[n] \oplus \sum_{e \in E} G_e'[e] \oplus G_g) .
\end{equation}

In the discussion below we use $\textrm{Core}(G_n, G_e, G_g)$ to denote the operations in Eq \ref{eq:edge_update}, \ref{eq:node_update}, and \ref{eq:global_update} above for brevity (Eq \ref{eq:gnn}):

\begin{equation}\label{eq:gnn}
G_n', G_e', G_g' = \textrm{Core}(G_n, G_e, G_g) .
\end{equation}

For more thorough introduction to graph neural networks, please refer to \cite{battaglia2018relational}.

\subsubsection*{Critic networks (Q-Value estimators)}

The two Q-value estimation networks have identical architecture. Given a state $s$ with features encoded in graph tuple $(G_n, G_e, G_g)$ and an action $a$, the Q-networks first concatenates the action (commanded joint velocities) with the corresponding robot node features, and the new node features are embedded into a higher dimensional embedding space using node embedding function $E_{n}$ to produce the node embedding inputs to the core GNN. The edge features and global features are also embedded into a higher dimensional embedding space using learned node embedding $E_{n}$ and edge embedding $E_{e}$ functions respectively (Eq \ref{eq:q_embedding}):

\begin{equation}\label{eq:q_embedding}
L_n, L_e, L_g = E_{n}(G_n \oplus a), E_{e}(G_e), E_{g}(G_g) .
\end{equation}

Then, the resulting graph embedding made up of the feature embeddings are passed through the core GNN for processing, and the output Q-value predicted from the global features of the output graph with a scalar prediction function $P_Q$ (Eq \ref{eq:q_output}):

\begin{equation}\label{eq:q_output}
Q(s, a) = P_Q(\textrm{Core}(L_n, L_e, L_g)_g) .
\end{equation}

\subsubsection*{Policy network}

The policy network uses a similar embedding process as the Q-networks, but without the action concatenation (Eq \ref{eq:pi_embedding}):

\begin{equation}\label{eq:pi_embedding}
L_n, L_e, L_g = E_{n}(G_n), E_{e}(G_e), E_{g}(G_g) .
\end{equation}

After passing through the core GNN, the output actions are predicted from the node features corresponding to the robot nodes $N_r$, using another prediction function $P_{\pi}$ (Eq \ref{eq:policy_output}):

\begin{equation}\label{eq:policy_output}
\pi(s) = P_{\pi}(\textrm{Core}(L_n, L_e, L_g)_n[n_r]) \quad \forall n_r \in N_r .
\end{equation}

All the learned functions (three embedding functions and three update functions) have the same architecture. Each is a fully connected network where each layer consist of a matrix multiplication followed by a layer normalization operation\cite{ba2016layer} and Gaussian Error Linear Unit (GELU) activation\cite{hendrycks2016gaussian}.

Both the value ($P_{Q}$) and policy ($P_{\pi}$) prediction networks are similar in architecture to the main learned update functions, but much smaller in size.

The experiments are conducted with JAX\cite{jax2018github}, with models built in the Jraph library\cite{jraph2020github} and the Flax library\cite{flax2020github}. Please refer to Supplementary Table S5 for the model shapes.

\subsection*{Training}

During training, 64 actors are used to continuously and asynchronously generate training episodes for insertion into a first-in-first-out replay buffer implemented using the Reverb library\cite{cassirer2021reverb}, updating their local copies of the policy network from the learner after each episode. The learner samples uniformly from the replay buffer. The replay buffer enforces a fixed sample-to-insert ratio by blocking insertion or sampling as necessary. The learner uses a single NVIDIA A100 GPU, and each actor uses an Intel Ice Lake Xeon CPU core. The training process takes $6$ to $24$ hours depending on problem size to achieve $95$\% task solution rate, and the quality of trajectories continues to improve (solving more tasks, reducing execution time, and reducing number of collisions) for an additional $2$ to $6$ days. Please refer to Supplementary Material Table S5 for the main hyper-parameters used in the experiments. These hyper-parameters were tuned with the Google Vizier\cite{vizier} black-box optimizer.




\clearpage

\bibliography{roboballet}

\bibliographystyle{Science}

\section*{ACKNOWLEDGEMENTS}
\textbf{Funding:} This work was funded by Google DeepMind and Intrinsic.\\
\textbf{Author contributions:} ML, KA conceptualized, designed, and conducted the experiments. ML, KA, and JS developed the core algorithmic framework. ML and KG developed and implemented the experimental infrastructure. ML, KA, KG, JS, and SS proposed improvements to the system. ML, KA, SS, and JS prepared and authored the paper. JS, KA, and ZL provided research guidance and planning. KA, JS, SS, and ZL edited and revised the paper. TK managed project and provided organizational support and funding.\\
\textbf{Competing interests:} Google DeepMind has submitted US and WO patent applications for the techniques described in this paper.\\
\textbf{Data and materials availability:} The open sourced version of the system described in this paper, which is implemented with MuJoCo physics simulator\cite{todorov2012mujoco}, is available at https://doi.org/10.5281/zenodo.16546403.

\section*{SUPPLEMENTARY MATERIALS}
\textbf{Video S1}: Rendered trajectory for the training setup (four robots).\\
\textbf{Video S2}: Rendered trajectory for the evaluation setup (four robots).\\
\textbf{Video S3}: Rendered trajectory for the training setup (eight robots).\\
\textbf{Video S4}: Rendered trajectory for the evaluation setup (eight robots).\\
\textbf{Table S5}: Experiment hyper-parameters.\\

\clearpage

\listoffigures

\figclearpage

\newgeometry{textwidth=7.3in,textheight=25cm,voffset=0.0cm}


\pagenumbering{gobble}


\begin{Movie}[H]
\centering
\includegraphics[width=7.3in]{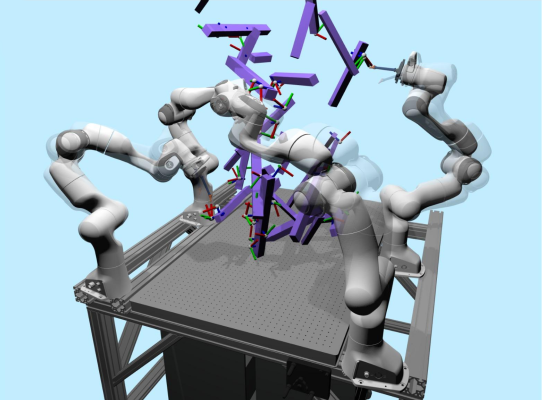}
\caption{\textbf{RoboBallet: Planning for Multi-Robot Reaching with Graph Neural Networks and Reinforcement Learning.}\\
We train the model on environments with randomized robot placement, randomly generated obstacles, and randomly generated tasks. These randomized environments are generated per episode in training. Throughout the training process the model interacts with about 1 million of these randomized environments. Each RGB coordinate frame on the obstacles represents an end-effector pose the robots need to reach. For testing, the evaluation setup with a fixed robot placement, real world obstacles, and pre-sampled tasks. Despite having never seen the evaluation robot placements, obstacles, or tasks during training, the model, trained on the randomized training environment, is able to transfer without any additional training to the evaluation cell, and generates time-efficient and collision-free trajectories.
\href{https://youtu.be/uqw7hTlk_BQ}{[Video Link (YouTube)]}
}
\label{fig:train_eval_setup}
\end{Movie}

\figclearpage

\begin{Movie}[H]

\includegraphics[width=7.3in]{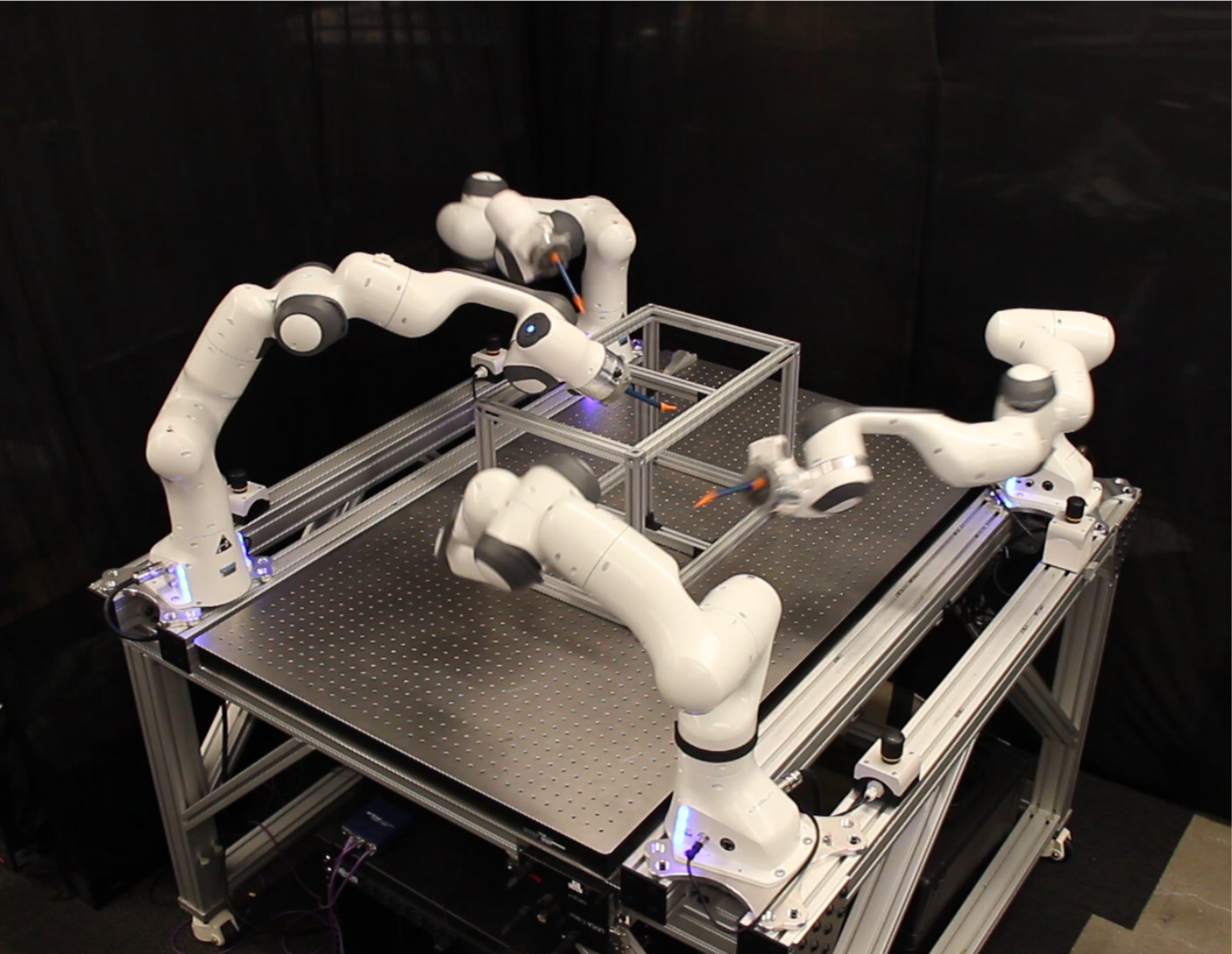}
\caption{\textbf{Four Franka Panda robots executing a plan generated by RoboBallet}. 
\href{https://youtu.be/JKymAhfFiRE}{[Video Link (YouTube)]}
}
\label{fig:real_robot_snapshots}
\end{Movie}

\figclearpage

\figclearpage

\figclearpage

\figclearpage

\figclearpage

\begin{figure}[H]
\centering
\includegraphics[width=12.7cm]{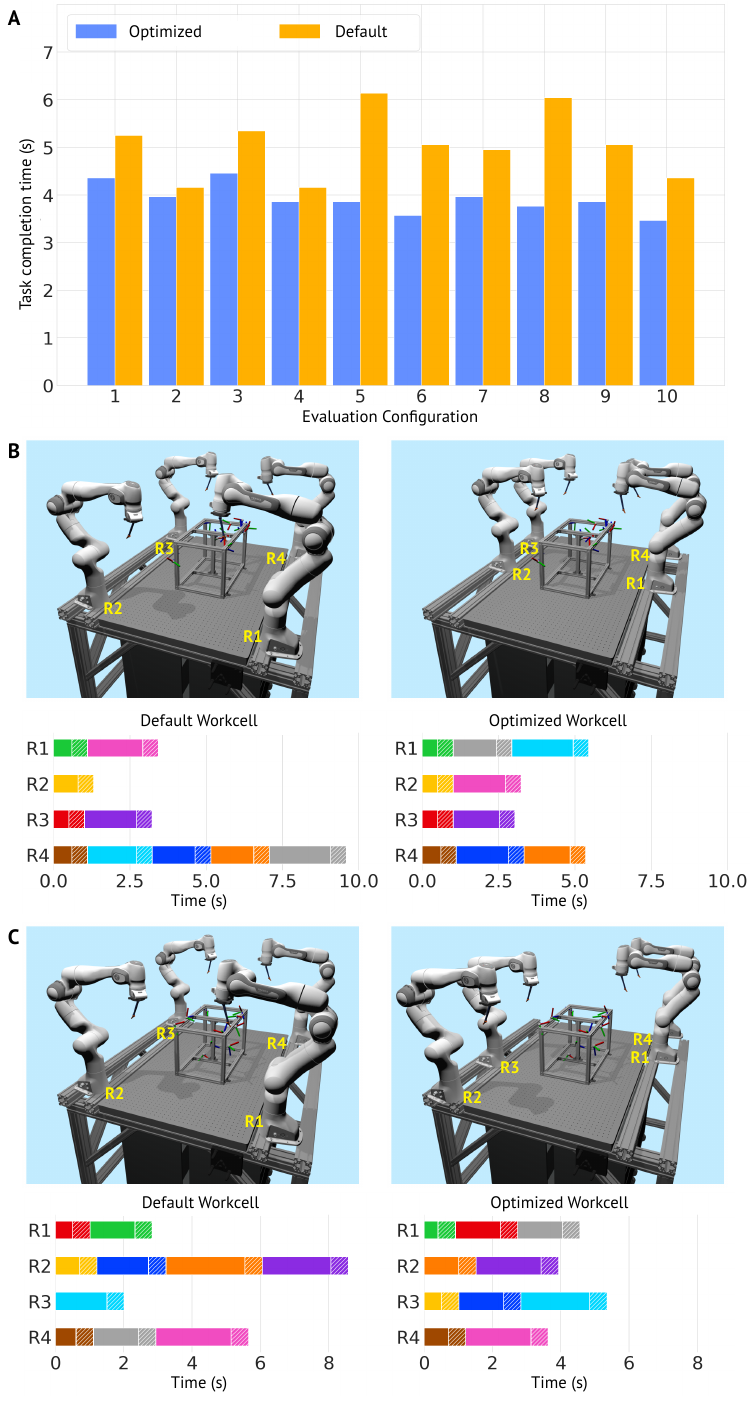}
\end{figure}

\figclearpage


\begin{figure}[H]
    \caption{
        \textbf{Layout optimization of the workcell.}\\
        (\textbf{A}) Execution time comparison between optimized and default layouts on the 10 evaluation configurations (each with 10 tasks). (\textbf{B}) Execution time breakdown of configuration \#5, the configuration with the biggest improvements. Each bar represents the tasks done by a robot in the trajectory, and each block represents the motion time and dwell time (simulating time needed to perform work at the site, e.g., a spot welding task). Each task has a unique color so that the re-distribution can be observed directly. (\textbf{C}) Same for configuration \#8.
    }
\label{fig:layout_optimization}
\end{figure}

\figclearpage

\clearpage

\setcounter{table}{4}
\renewcommand{\thetable}{S\arabic{table}}

\begin{table}[ht]
\begin{center}
\begin{tabular}{|c|c|c|}
    \hline
    Param & Value & Description \\
    \hline
    $T_{max}$ & $10 \times 10^{6}$ & Maximum number of training steps \\
    \hline
    $B$ & $128$ & Training batch size (number of ($s$, $a$, $r$, $s'$) transitions) \\
    \hline
    $\eta_{\pi}$ & $5 \times 10^{-5}$ & Policy learning rate \\
    \hline
    $\eta_{Q}$ & $5 \times 10^{-5}$ & Twin critic learning rate \\
    \hline
    $\lambda_{\eta_{\pi}}$ & $0.98$ & Policy learning rate decay rate (per 100,000 steps) \\
    \hline
    $\lambda_{\eta_{Q}}$ & $0.95$ & Twin critic learning rate decay rate (per 100,000 steps) \\
    \hline
    $\sigma_{t}$ & $5 \times 10^{-4}$ & Q-network target noise standard deviation \\
    \hline
    $\tau_{\pi}$ & $4 \times 10^{-5}$ & Policy target network update delay \\
    \hline
    $\tau_{Q}$ & $8 \times 10^{-5}$ & Twin critic target network update delay \\
    \hline
    $\gamma$ & $0.94$ & Future reward discount factor \\
    \hline
    $E_{n}, E_{g}$ & $\textrm{MLP}(512, 7)$ & Node and global embedding functions network architecture \\
    \hline
    $E_{e}$ & $\textrm{MLP}(256, 6)$ & Edge embedding function network architecture \\
    \hline
    $U_{n}, U_{g}$ & $\textrm{MLP}(512, 7)$ & Node and global update functions network architecture \\
    \hline
    $U_{e}$ & $\textrm{MLP}(256, 6)$ & Edge update function network architecture \\
    \hline
    $P_{\pi}, P_{Q}$ & $\textrm{MLP}(64, 2)$ & Action prediction network architecture \\
    \hline
    $N_{replay}$ & $ \frac{2 \times 10^{7}}{N_{tasks}} $ & Maximum replay buffer size \\
    \hline
    $S_{replay}$ & $ 5.0 $ & Replay sample-to-insert ratio \\
    \hline
    $t_{s}$ & $ 0.1 $ & Environment time-step size (seconds) \\
    \hline
    $C_{col}$ & $ 15.0 $ & Collision penalty coefficient \\
    \hline
\end{tabular}
\caption{Table of training hyper-parameters, where $\textrm{MLP}(n, m)$ denotes a fully connected multi-layer perceptron (MLP) network of $m$ layers each $n$ nodes wide.}
\label{table:hparams}
\end{center}
\end{table}

\end{document}